\documentclass[runningheads]{llncs}
\usepackage[T1]{fontenc}
\usepackage{graphicx}
\usepackage{amsmath,amssymb,bm}
\usepackage{booktabs}
\usepackage{multirow}
\usepackage{array}
\usepackage{xcolor}
\usepackage{hyperref}
\usepackage{url}
\usepackage{marvosym}

\newcommand{\gray}[1]{\textcolor{gray}{#1}}

\newcommand{\method}{CARE}
\begin{document}
\title{CARE: Constrained Attention Refinement for Fine‑Grained Visual Classification via Teacher‑Student Distillation}
\titlerunning{CARE}

\author{
    Ruibo Wen\inst{1}\orcidID{0009-0008-1671-2279} \and
    Hang Shao\inst{1}\orcidID{0000-0002-1322-4789} \and \\
    Yiming Lei\inst{1}\orcidID{0000-0002-1349-7074}\textsuperscript{\Letter}
}
\authorrunning{R. Wen et al.}
%
\institute{
College of Computer Science and Technology, Qingdao University, Qingdao, China\\
\email{leiyiming@qdu.edu.cn}
}
%
\maketitle              
\begingroup
\renewcommand{\thefootnote}{}
\footnotetext{\Letter\ Corresponding author.}
\endgroup

\begin{abstract}
Fine-grained visual classification requires models to recognize subtle local traits while exposing the visual evidence behind their predictions. Class-specific attention pathways provide a natural basis for interpretable recognition, but their constrained prediction structure limits discriminative capacity and underuses intermediate representations from strong pretrained backbones. To address this problem, we propose \method{}, a constrained attention refinement framework for interpretable fine-grained recognition via teacher-student distillation. CARE keeps the final prediction and explanation within a class-specific attention student, while introducing a training-only auxiliary query teacher that reads selected intermediate DINOv2 layers with learnable queries. The teacher fuses multi-level representations and transfers logit-standardized class-discriminative knowledge to the student. To further refine the explanation pathway, we design diversity and sparsity terms to regularize student attention heads, reducing redundancy and encouraging compact trait localization. Experiments on CUB, Oxford-IIIT Pet, Stanford Dogs, and Stanford Cars show that CARE achieves strong classification performance under an interpretable frozen-backbone setting, reaching 78.5\% Top-1 accuracy on CUB. Faithfulness analysis with insertion and deletion metrics further indicates that the top-ranked attention regions retain class-relevant evidence for explanation. The source code is publicly available at \url{https://github.com/panpan0814/CARE}.
\end{abstract}

\section{Introduction}

Fine-grained visual classification (FGVC) aims to separate categories that differ in subtle local cues, including species, breeds, and object subtypes. In biodiversity monitoring, industrial inspection, and medical image analysis, a small visual detail may determine the correct category. Models used in these settings therefore need to expose the visual evidence behind their predictions, rather than rely on background bias or holistic shortcuts~\cite{bilinear2015,transfg2022}.

Vision Transformers (ViTs) are strong backbones for FGVC because their token interactions capture both local parts and global context~\cite{vit2021,swin2021,transfg2022}. Self-supervised models such as DINO and DINOv2 further provide transferable frozen representations for downstream recognition tasks~\cite{dino2021,dinov22023}. Post-hoc attribution methods can explain decisions after a predictive model has been trained, but they do not require the classifier to use trait-level evidence when making predictions~\cite{gradcam2017,attentionrollout2020,transformerinterpretability2021}. Intrinsically interpretable models connect prediction and explanation more directly, although they often introduce additional structures that are not always suitable for lightweight adaptation of frozen ViT backbones~\cite{intr2024,prototree2021}.

Visual Prompt Tuning (VPT) adapts frozen ViTs with learnable visual prompt tokens and provides a parameter-efficient alternative to full fine-tuning~\cite{vpt2022}. Its focus, however, is mainly on recognition performance, and it does not provide class-specific visual explanations. Prompt-CAM extends prompt-based adaptation to intrinsic interpretability by using class-specific prompt outputs for prediction, with prompt-to-patch attention maps serving as class-level explanations~\cite{promptcam2025}. This design is efficient and compatible with frozen ViTs, but prediction is still limited to final-layer prompt representations and a shared scoring head. As a result, it may not fully use intermediate Transformer representations that contain complementary local, part-level, and semantic evidence.

We propose \method{}, a constrained attention refinement framework for interpretable fine-grained recognition. CARE keeps a class-specific prompt-attention student as the final prediction and explanation pathway, and improves it through training supervision. Diversity and sparsity losses regularize the student attention heads to reduce redundancy and encourage compact trait localization. A training-only auxiliary query teacher reads selected intermediate DINOv2 layers and transfers multi-level discriminative knowledge to the student through logit-standardized knowledge distillation (KD)~\cite{kd2015,lskd2024}. During inference, the teacher branch is removed, so prediction and explanation are produced by the student alone.
Our contributions are summarized as follows:
\begin{itemize}
    \item We formulate constrained attention refinement for interpretable FGVC, aiming to improve discriminative capacity while keeping prediction and explanation within a class-specific attention pathway.
    \item We introduce a training-only auxiliary query teacher that reads multi-level DINOv2 representations and distills logit-standardized class-discriminative knowledge to the student.
    \item We apply diversity and sparsity regularization to student attention heads, reducing redundancy and encouraging compact localization of fine-grained traits.
    \item We evaluate CARE on CUB, Oxford-IIIT Pet, Stanford Dogs, and Stanford Cars using classification accuracy, insertion/deletion faithfulness, component ablations, and teacher-layer analysis.
\end{itemize}

\section{Related Work}

\paragraph{Fine-grained visual classification.}
FGVC distinguishes visually similar categories using subtle local traits. Early methods relied on part localization, bilinear interactions, or region modeling~\cite{part_rcnn2014,bilinear2015}. Transformers reduce the need for hand-crafted part modeling by representing images as patch tokens~\cite{vit2021,deit2021,swin2021}, and TransFG adapts ViTs to FGVC through discriminative token selection~\cite{transfg2022}. Furthermore, self-supervised foundation models such as DINO and DINOv2 provide strong frozen representations for downstream recognition~\cite{dino2021,dinov22023}. 
Our work studies how to exploit multi-level DINOv2 representations while preserving class-specific visual evidence for interpretable fine-grained recognition.

\paragraph{Prompt learning and parameter-efficient adaptation.}
Prompt learning adapts pretrained models with a small number of trainable parameters. VPT introduces visual prompts into frozen ViTs~\cite{vpt2022}, while CoOp and MaPLe adapt vision-language models with learned prompts~\cite{coop2022,maple2023}. Recent work on PEFT further improves prompt-based adaptation~\cite{davpt2025}. These methods mainly target recognition performance rather than intrinsic explanation. Prompt-CAM extends prompt-based adaptation by using class-specific prompts for both prediction and explanation~\cite{promptcam2025}. In contrast, \method{} focuses on refining the class-specific attention pathway through attention regularization and training-only teacher-student distillation.

\paragraph{Visual attribution and interpretability.}
Post-hoc methods such as Grad-CAM, Attention Rollout, and Transformer attribution explain trained classifiers after the decision function has been learned~\cite{gradcam2017,attentionrollout2020,transformerinterpretability2021}. They are useful diagnostic tools, but they do not require the classifier itself to rely on interpretable visual evidence. Intrinsic methods instead build explanation structures into the model. MCTformer uses class tokens for localization~\cite{mctformer2022}, ProtoTree and ProtoPFormer use prototype reasoning~\cite{prototree2021,protopformer2024}, and INTR uses class-specific queries in an encoder--decoder architecture~\cite{intr2024}. Class-specific prompt-attention models offer a simpler and more parameter-efficient form of intrinsic explanation, but their constrained prediction pathway can limit discriminative capacity. \method{} addresses this limitation by preserving the attention-based explanation pathway while strengthening it with multi-level teacher supervision.

\paragraph{Knowledge distillation and intermediate representations.}
Knowledge distillation transfers softened outputs or intermediate features from a teacher to a student~\cite{kd2015}. Decoupled distillation and logit standardization further refine the transferred signal~\cite{dkd2022,lskd2024}. In ViTs, intermediate layers encode complementary cues, ranging from local patterns to semantic abstractions. Visual Query Tuning shows that learnable queries can access such representations efficiently~\cite{vqt2023}. Inspired by this idea, the Auxiliary Query Teacher in \method{} reads selected DINOv2 layers, produces teacher logits, and transfers multi-level discriminative knowledge to the Constrained Attention Student only during training.

\section{Methodology}

\begin{figure}[t]
    \centering
    \includegraphics[width=\textwidth]{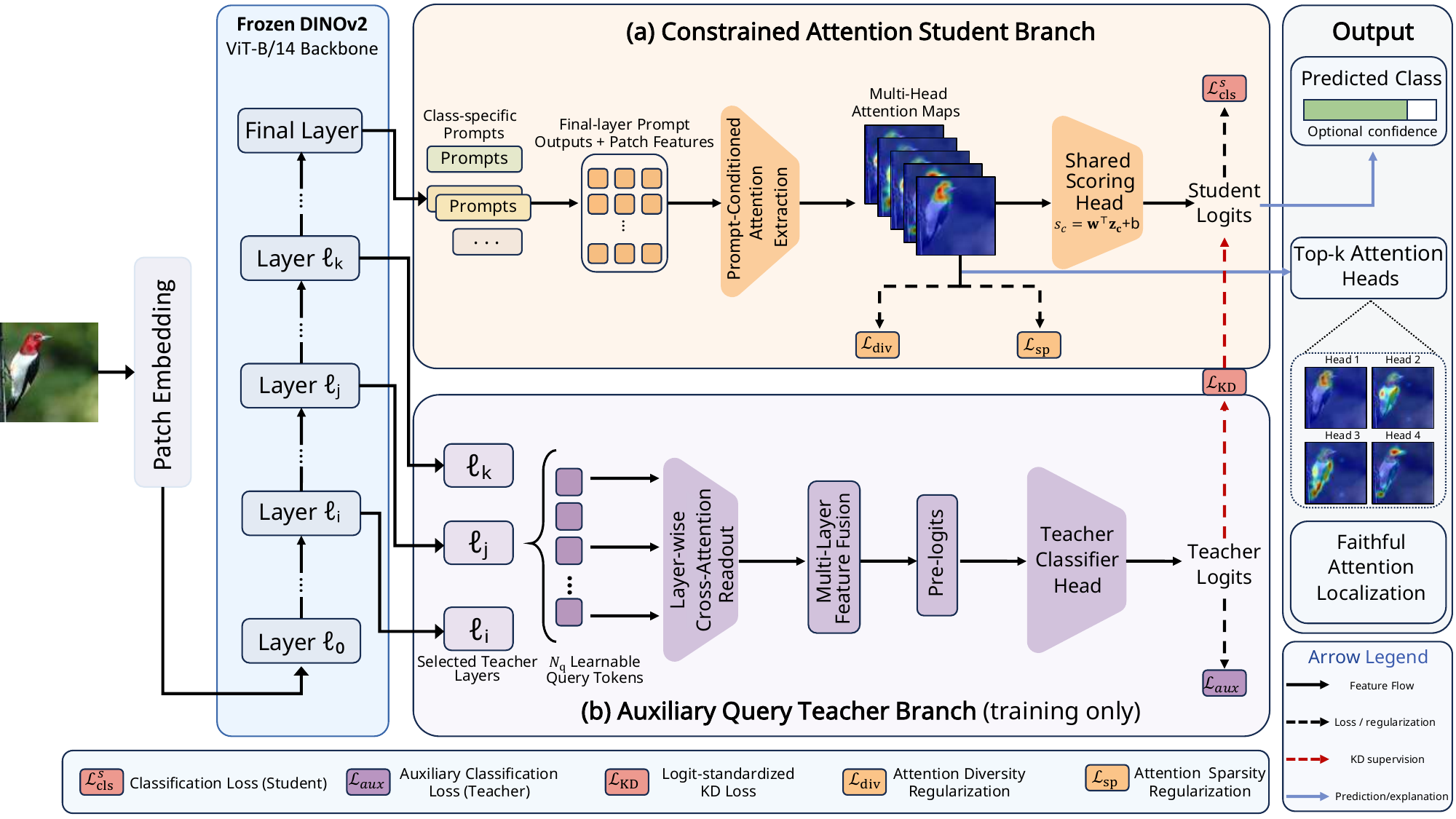}
    \caption{
    Overall architecture of \method{}.
    The framework consists of (a) a Constrained Attention Student branch for prediction and explanation, and (b) a training-only Auxiliary Query Teacher branch for multi-layer distillation.
    }
    \label{fig:framework}
\end{figure}

\subsection{Overview}

Given an input image $\mathbf{x}$ and label $y\in\{1,\ldots,C\}$, our goal is to strengthen the discriminative capacity of a class-specific attention pathway while preserving it as the final prediction and explanation source. Fig.~\ref{fig:framework} shows the proposed \method{} framework, which is built on a frozen DINOv2 Vision Transformer~\cite{dinov22023}. During training, \method{} contains two branches: (a) a Constrained Attention Student for prediction and explanation, and (b) a training-only Auxiliary Query Teacher for multi-layer distillation.

The Constrained Attention Student follows a class-specific prompt-attention design for prediction and explanation. Class-specific prompt outputs are scored by a shared vector to produce logits, and the corresponding prompt-to-patch multi-head attention maps are used as class-specific explanations. We regularize these maps with diversity and sparsity losses to reduce redundancy across heads and encourage more compact localization. This keeps the explanation source tied to the same class-specific pathway used for prediction. The Auxiliary Query Teacher is introduced to exploit hierarchical evidence that may not be fully represented by the final student prompt outputs. It reads features from selected intermediate layers, where cues such as local structure, part boundaries, texture, and higher-level semantics are useful for fine-grained recognition.

The frozen backbone extracts both intermediate and final-layer features. The student produces logits and attention maps through the final class-specific attention pathway. In parallel, the teacher uses learnable query tokens to read selected intermediate layers, fuses the query-conditioned readouts, and predicts teacher logits. These logits supervise the student through logit-standardized KD~\cite{kd2015,lskd2024}. At inference, the teacher branch is removed, and the student alone provides both prediction and explanation.

\subsection{Constrained Attention Student Branch}

The student branch follows the class-specific prompt-attention principle of Prompt-CAM~\cite{promptcam2025}. Instead of using a generic CLS-token classifier, class-specific prompt tokens are used to compute category scores from their corresponding prompt outputs. The same prompts also query image patches through attention, so the prompt-to-patch attention maps can be used as class-specific visual evidence.

Let $F_{\theta}$ denote the frozen DINOv2 ViT with $N$ Transformer layers. The input image $\mathbf{x}$ is mapped to patch tokens $\mathbf{E}_{0}\in\mathbb{R}^{M\times D}$, where $M$ is the number of image patches and $D$ is the hidden dimension. We introduce class-specific prompt tokens $\mathbf{P}=[\mathbf{p}^{1},\ldots,\mathbf{p}^{C}]^{\top}\in\mathbb{R}^{C\times D}$, where $\mathbf{p}^{c}\in\mathbb{R}^{D}$ is the learnable prompt token for class $c$. After the final Transformer layer, the class prompt outputs are $\mathbf{Z}_{N}^{s}=[\mathbf{z}_{N}^{s,1},\ldots,\mathbf{z}_{N}^{s,C}]^{\top}\in\mathbb{R}^{C\times D}$.

A shared scoring vector $\mathbf{w}_{s}\in\mathbb{R}^{D}$ and bias $b_s\in\mathbb{R}$ map each class prompt to a scalar logit:
\begin{equation}
    s_{c}^{s}
=
\mathbf{w}_{s}^{\top}\mathbf{z}_{N}^{s,c}
+
b_s,
\qquad c=1,\ldots,C.
\end{equation}
The student logit vector and classification loss are
\begin{equation}
    \mathbf{s}^{s}=[s_{1}^{s},\ldots,s_{C}^{s}]^{\top}\in\mathbb{R}^{C},\qquad
    \mathcal{L}_{\mathrm{cls}}^{s}
    =
    -\log
    \frac{\exp(s_{y}^{s})}
    {\sum_{c=1}^{C}\exp(s_{c}^{s})}.
\end{equation}

For class $c$ and attention head $r$, let $\boldsymbol{\alpha}^{c,r}\in\mathbb{R}^{M}$ denote the final-layer prompt-to-patch attention map over image patches. With $R$ attention heads, the attention-map set for class $c$ is $\mathcal{A}^{c}=\{\boldsymbol{\alpha}^{c,r}\}_{r=1}^{R}$. The Constrained Attention Student keeps the class-specific prediction pathway and adds $\mathcal{L}_{\mathrm{div}}$ and $\mathcal{L}_{\mathrm{sp}}$ to regularize the student attention maps.

\subsection{Auxiliary Query Teacher Branch}

The Auxiliary Query Teacher reads features from selected layers of the frozen backbone. Let the frozen backbone contain $N$ Transformer layers indexed as $\ell_0,\ell_1,\ldots,\ell_{N-1}$. The selected teacher-layer set is denoted as $\mathcal{S}=\{\ell_i,\ell_j,\ell_k\}$, where $0\leq i<j<k\leq N-1$. Unless otherwise stated, we use the zero-based DINOv2 layer indices $\{4,8,11\}$.

For layer $\ell\in\mathcal{S}$, the context tokens are $\mathbf{H}_{\ell}=[\mathbf{h}_{\ell}^{1},\ldots,\mathbf{h}_{\ell}^{M_{\ell}}]^{\top}\in\mathbb{R}^{M_{\ell}\times D}$, where $M_{\ell}$ denotes the number of context tokens available to the teacher branch. These tokens contain the intermediate non-prompt representations produced by the frozen backbone.

We introduce $N_q$ learnable query tokens $\mathbf{U}_{\ell}\in\mathbb{R}^{N_q\times D}$ for each selected layer, with $N_q=8$ in the main configuration. The query, key, and value matrices are
\begin{equation}
\begin{aligned}
    \mathbf{Q}_{\ell}&=\mathbf{U}_{\ell}\mathbf{W}_{\ell}^{Q},&
    \mathbf{K}_{\ell}&=\mathbf{H}_{\ell}\mathbf{W}_{\ell}^{K},&
    \mathbf{V}_{\ell}&=\mathbf{H}_{\ell}\mathbf{W}_{\ell}^{V},
\end{aligned}
\end{equation}
where $\mathbf{W}_{\ell}^{Q},\mathbf{W}_{\ell}^{K},\mathbf{W}_{\ell}^{V}\in\mathbb{R}^{D\times D}$. The layer-wise cross-attention matrix is
\begin{equation}
    \mathbf{A}_{\ell}
    =
    \operatorname{softmax}_{\mathrm{ctx}}
    \left(
    \frac{\mathbf{Q}_{\ell}\mathbf{K}_{\ell}^{\top}}{\sqrt{D}}
    \right)
    \in\mathbb{R}^{N_q\times M_{\ell}},
\end{equation}
where the softmax is applied over the context-token dimension. Equivalently,
\begin{equation}
    A_{\ell,q,m}
    =
    \frac{
    \exp(\langle \mathbf{Q}_{\ell}^{q},\mathbf{K}_{\ell}^{m}\rangle/\sqrt{D})
    }
    {
    \sum_{m'=1}^{M_{\ell}}\exp(\langle \mathbf{Q}_{\ell}^{q},\mathbf{K}_{\ell}^{m'}\rangle/\sqrt{D})
    }.
\end{equation}
The query-conditioned readout is
\begin{equation}
    \mathbf{R}_{\ell}=\mathbf{A}_{\ell}\mathbf{V}_{\ell}\in\mathbb{R}^{N_q\times D}.
\end{equation}
In implementation, this readout is realized as a residual multi-head cross-attention module followed by a lightweight feed-forward network, and the readout dimension is kept the same as the backbone embedding dimension.

We average the query readouts and fuse the selected layers:
\begin{equation}
    \rho(\mathbf{R}_{\ell})
    =
    \frac{1}{N_q}
    \sum_{q=1}^{N_q}
    \mathbf{R}_{\ell}^{q},
    \qquad
    \mathbf{g}
    =
    \phi
    \left(
    \operatorname{Concat}
    [\rho(\mathbf{R}_{\ell_i}),\rho(\mathbf{R}_{\ell_j}),\rho(\mathbf{R}_{\ell_k})]
    \right),
\end{equation}
where $\rho(\mathbf{R}_{\ell})\in\mathbb{R}^{D}$,
$\phi:\mathbb{R}^{3D}\rightarrow\mathbb{R}^{D}$ is a lightweight fusion MLP,
and $\mathbf{g}\in\mathbb{R}^{D}$ is the fused teacher representation.

The teacher logits and auxiliary loss are
\begin{equation}
    \mathbf{s}^{t}=\mathbf{W}_{t}\mathbf{g}+\mathbf{b}_{t},\qquad
    \mathcal{L}_{\mathrm{aux}}
    =
    -\log
    \frac{\exp(s_y^t)}
    {\sum_{c=1}^{C}\exp(s_c^t)}.
\end{equation}
Here, $\mathbf{W}_{t}\in\mathbb{R}^{C\times D}$, $\mathbf{b}_{t}\in\mathbb{R}^{C}$, and $\mathbf{s}^{t}\in\mathbb{R}^{C}$.
The teacher branch is used only for training-time supervision and is not used as the explanation source.

\subsection{Logit-Standardized Knowledge Distillation}

Teacher and student logits may have different scales, so we standardize them before distillation~\cite{lskd2024}. For $\mathbf{s}\in\mathbb{R}^{C}$,
\begin{equation}
    \widehat{\mathbf{s}}
    =
    \frac{\mathbf{s}-\mu(\mathbf{s})\mathbf{1}}
    {\sigma(\mathbf{s})+\epsilon},
    \quad
    \mu(\mathbf{s})=\frac{1}{C}\sum_{j=1}^{C}s_j,
    \quad
    \sigma(\mathbf{s})=
    \left(
    \frac{1}{C}\sum_{j=1}^{C}(s_j-\mu(\mathbf{s}))^2
    \right)^{1/2}.
\end{equation}
Here, $\mathbf{1}\in\mathbb{R}^{C}$ is an all-one vector and $\epsilon$ is a small constant for numerical stability.
The standardized teacher and student distributions are
\begin{equation}
    \mathbf{q}^{t}
    =
    \operatorname{softmax}
    \left(
    \frac{\operatorname{sg}(\widehat{\mathbf{s}}^{t})}{T}
    \right),
    \qquad
    \mathbf{q}^{s}
    =
    \operatorname{softmax}
    \left(
    \frac{\widehat{\mathbf{s}}^{s}}{T}
    \right),
\end{equation}
where $T>0$ is the distillation temperature, $\mathbf{q}^{t},\mathbf{q}^{s}\in\mathbb{R}^{C}$ are probability distributions, and $\operatorname{sg}(\cdot)$ denotes stop-gradient. Thus, the KD term transfers knowledge from the teacher to the student without updating the teacher through the distillation loss.

The KL divergence and KD loss are defined as:
\begin{equation}
    \operatorname{KL}(\mathbf{q}^{t}\,\|\,\mathbf{q}^{s})
    =
    \sum_{c=1}^{C}
    q_c^t
    \log
    \left(
    \frac{q_c^t+\epsilon}{q_c^s+\epsilon}
    \right),
    \qquad
    \mathcal{L}_{\mathrm{KD}}
    =
    T^{2}
    \operatorname{KL}
    (\mathbf{q}^{t}\,\|\,\mathbf{q}^{s}).
\end{equation}
This loss transfers the teacher's relative class distribution to the Constrained Attention Student.

\subsection{Attention Diversity and Sparsity Regularization}
\label{sec:attention_reg}

For the ground-truth class $y$, the normalized attention map of head $r$ is
\begin{equation}
    \widetilde{\boldsymbol{\alpha}}^{y,r}
    =
    \frac{\boldsymbol{\alpha}^{y,r}}
    {\sum_{m=1}^{M}\alpha_m^{y,r}+\epsilon}
    \in\mathbb{R}^{M},
\end{equation}
where $\alpha_m^{y,r}$ denotes the attention value on patch $m$ and $\epsilon$ is a small constant.

The diversity loss is the average pairwise cosine similarity between different heads:
\begin{equation}
    \mathcal{L}_{\mathrm{div}}
    =
    \frac{1}{R(R-1)}
    \sum_{\substack{r,r'=1\\r\neq r'}}^{R}
    \frac{
    (\widetilde{\boldsymbol{\alpha}}^{y,r})^{\top}
    \widetilde{\boldsymbol{\alpha}}^{y,r'}
    }
    {
    \|\widetilde{\boldsymbol{\alpha}}^{y,r}\|_2
    \|\widetilde{\boldsymbol{\alpha}}^{y,r'}\|_2
    +
    \epsilon
    }.
\end{equation}

For sparsity, we first select the top-$k$ attention heads of the ground-truth class and then regularize their entropy.
The head-selection score is defined by the maximum prompt-to-patch attention response:
\begin{equation}
    \eta_r^y
    =
    \max_{m\in\{1,\ldots,M\}}
    \alpha_m^{y,r}.
\end{equation}
The selected head index set is
\begin{equation}
    \mathcal{I}_k^y
    =
    \operatorname{TopK}_{r\in\{1,\ldots,R\}}
    \left(
    \operatorname{sg}(\eta_r^y), k
    \right),
\end{equation}
where $\operatorname{sg}(\cdot)$ denotes stop-gradient.
The discrete top-$k$ operation determines which heads are regularized, while gradients are back-propagated through the selected attention maps.

The sparsity loss is the normalized entropy of the selected heads:
\begin{equation}
    \mathcal{L}_{\mathrm{sp}}
    =
    \frac{1}{|\mathcal{I}_k^y|}
    \sum_{r\in\mathcal{I}_k^y}
    \frac{
    \mathcal{H}
    \left(
    \widetilde{\boldsymbol{\alpha}}^{y,r}
    \right)
    }
    {\log M},
\end{equation}
where
\begin{equation}
    \mathcal{H}
    \left(
    \widetilde{\boldsymbol{\alpha}}^{y,r}
    \right)
    =
    -
    \sum_{m=1}^{M}
    \widetilde{\alpha}_m^{y,r}
    \log
    \left(
    \widetilde{\alpha}_m^{y,r}
    +
    \epsilon
    \right).
\end{equation}
Both regularizers act directly on the Constrained Attention Student attention maps.

\subsection{Overall Objective and Training}

The total objective is
\begin{equation}
\begin{aligned}
    \mathcal{L}_{\mathrm{total}}
    =
    \;
    \mathcal{L}_{\mathrm{cls}}^{s}
    +
    \lambda_{\mathrm{aux}}\mathcal{L}_{\mathrm{aux}}
    +
    \lambda_{\mathrm{KD}}\mathcal{L}_{\mathrm{KD}} 
    +
    \lambda_{\mathrm{div}}\mathcal{L}_{\mathrm{div}}
    +
    \lambda_{\mathrm{sp}}\mathcal{L}_{\mathrm{sp}}.
\end{aligned}
\end{equation}
Here, $\lambda_{\mathrm{aux}}$, $\lambda_{\mathrm{KD}}$, $\lambda_{\mathrm{div}}$, and $\lambda_{\mathrm{sp}}$ are scalar loss weights.
The attention regularizers do not introduce additional trainable parameters; they only impose constraints on student attention maps.
The trainable parameters include the class-specific prompt tokens, the shared student scoring head, and the Auxiliary Query Teacher branch, while the DINOv2 backbone remains frozen.
During inference, the final prediction and explanation are produced by the Constrained Attention Student branch, and the Auxiliary Query Teacher is removed.

\section{Experiments}
\subsection{Datasets and Implementation Details}

\subsubsection{Datasets}
We evaluate \method{} on CUB-200-2011~\cite{cub2011}, Stanford Dogs~\cite{dog2011}, Oxford-IIIT Pet~\cite{pet2012}, and Stanford Cars~\cite{cars2013}. CUB is used for ablation studies, teacher-layer selection, and hyperparameter analysis. The remaining datasets test whether the observed behavior transfers across different fine-grained domains.

\subsubsection{Experimental settings}
All prompt-based models use a frozen DINOv2 ViT-B/14 backbone~\cite{dinov22023} with an input resolution of $224\times224$. Unless otherwise stated, models are trained for 60 epochs with SGD, momentum 0.9, weight decay 0.001, batch size 64, learning rate 0.005, and 10 warm-up epochs. For \method{}, we use zero-based teacher-layer indices $\{4,8,11\}$, $N_q=8$, $\lambda_{\mathrm{aux}}=0.5$, $\lambda_{\mathrm{KD}}=1.0$, $T=2.0$, $\lambda_{\mathrm{div}}=0.05$, $\lambda_{\mathrm{sp}}=0.01$, and $k=4$ for sparsity regularization.

\subsubsection{Evaluation Metrics}
Classification performance is measured by Top-1 accuracy. For explanation faithfulness, we report insertion and deletion scores. Class-specific prompt-attention methods are evaluated using their intrinsic attention maps, while Grad-CAM, Layer-CAM, and Attention Rollout are evaluated on the DINOv2 Linear Probing classifier~\cite{gradcam2017,layercam2021,attentionrollout2020}.

\subsubsection{Baselines}
We compare \method{} with both recognition-oriented and interpretable baselines. DINOv2 Linear Probing is included as a recognition-oriented reference with a frozen ViT-B/14 backbone and a linear head. For interpretable fine-grained recognition, we compare with INTR~\cite{intr2024}, ProtoTree~\cite{prototree2021}, TesNet~\cite{tesnet2022}, and the class-specific prompt-attention baseline Prompt-CAM~\cite{promptcam2025}. Unless otherwise stated, Linear Probing, Prompt-CAM, and \method{} are evaluated under the same frozen DINOv2 setting, while the remaining interpretable baseline results follow the corresponding published reports. Prompt-CAM is the closest baseline, while \method{} further introduces constrained attention refinement with training-only teacher-student distillation.

\subsection{Main Results}

The classification results show that \method{} consistently improves the class-specific prompt-attention baseline across all four fine-grained benchmarks, as reported in Table~\ref{tab:classification_accuracy}. Compared with Prompt-CAM, \method{} increases Top-1 accuracy from 74.1\% to 78.5\% on CUB, from 81.3\% to 81.9\% on Stanford Dogs, from 92.7\% to 93.6\% on Oxford-IIIT Pet, and from 84.3\% to 87.1\% on Stanford Cars.

These gains suggest that the auxiliary query teacher and logit-standardized distillation improve the discriminative ability of the class-specific attention pathway while preserving its intrinsic explanation mechanism.

\begin{table}[t]
\centering
\caption{
Top-1 classification accuracy (\%) on four fine-grained recognition benchmarks.
Gray values denote recognition-oriented Linear Probing results provided for reference.
Bold indicates the best result among the compared interpretable methods.
}
\label{tab:classification_accuracy}
\begin{tabular}{lcccccc}
\hline
Dataset & Linear Probing & INTR & ProtoTree & TesNet & Prompt-CAM & \method{} \\
\hline
CUB  & \gray{89.6} & 71.8 & 68.5 & 77.7 & 74.1 & \textbf{78.5} \\
Dog  & \gray{87.3} & 72.5 & 68.4 & 76.5 & 81.3 & \textbf{81.9} \\
Pet  & \gray{96.2} & 90.4 & --   & --   & 92.7 & \textbf{93.6} \\
Cars & \gray{88.2} & 86.8 & 70.0 & 84.6 & 84.3 & \textbf{87.1} \\
\hline
\end{tabular}
\end{table}

\subsection{Faithfulness Analysis}

Explanation faithfulness is evaluated with insertion and deletion scores under the two settings defined above: post-hoc explanations on the DINOv2 Linear Probing classifier and intrinsic attention maps from class-specific prompt-attention methods.
All methods follow the same perturbation protocol.
Saliency maps are upsampled to the image resolution, pixels are ranked by descending saliency, and each score is computed as the area under the target-class confidence curve over 50 perturbation steps.
The quantitative results are reported in Table~\ref{tab:faithfulness_combined}.

\begin{figure*}[!t]
    \centering
    \begin{minipage}{0.325\textwidth}\centering
        \includegraphics[width=\linewidth]{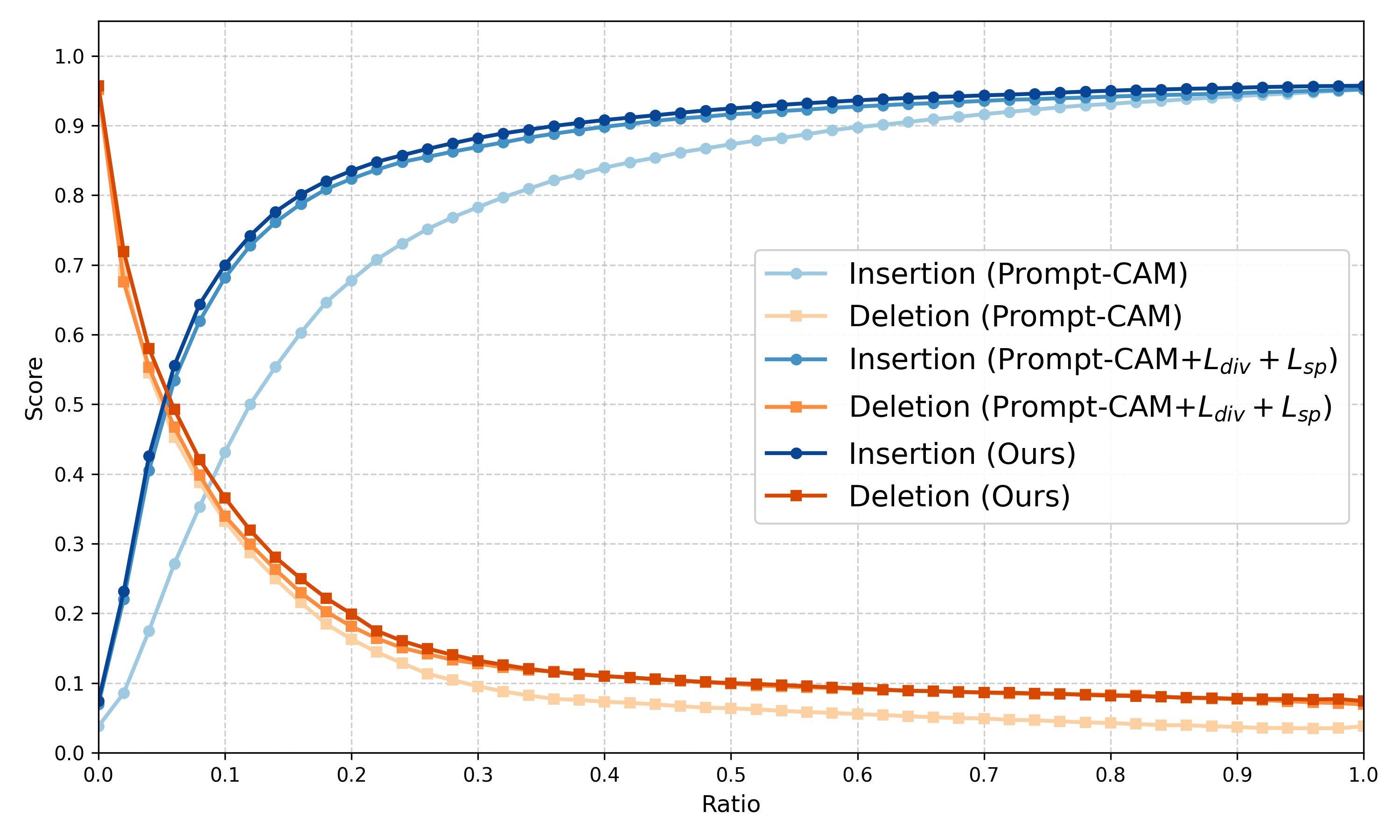}
        \vspace{-0.4em}\centerline{\small (a) CUB}
    \end{minipage}\hfill
    \begin{minipage}{0.325\textwidth}\centering
        \includegraphics[width=\linewidth]{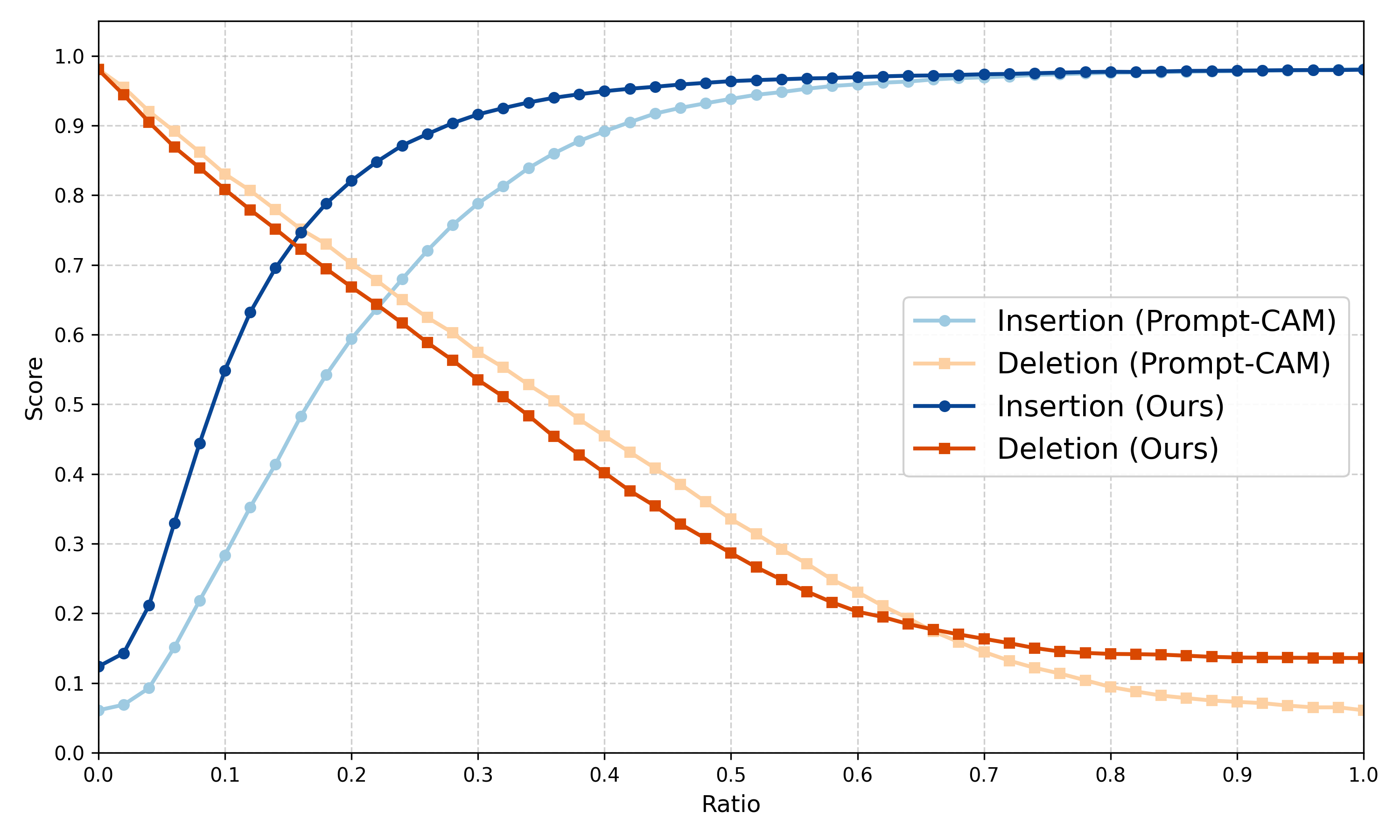}
        \vspace{-0.4em}\centerline{\small (b) Oxford-IIIT Pet}
    \end{minipage}\hfill
    \begin{minipage}{0.325\textwidth}\centering
        \includegraphics[width=\linewidth]{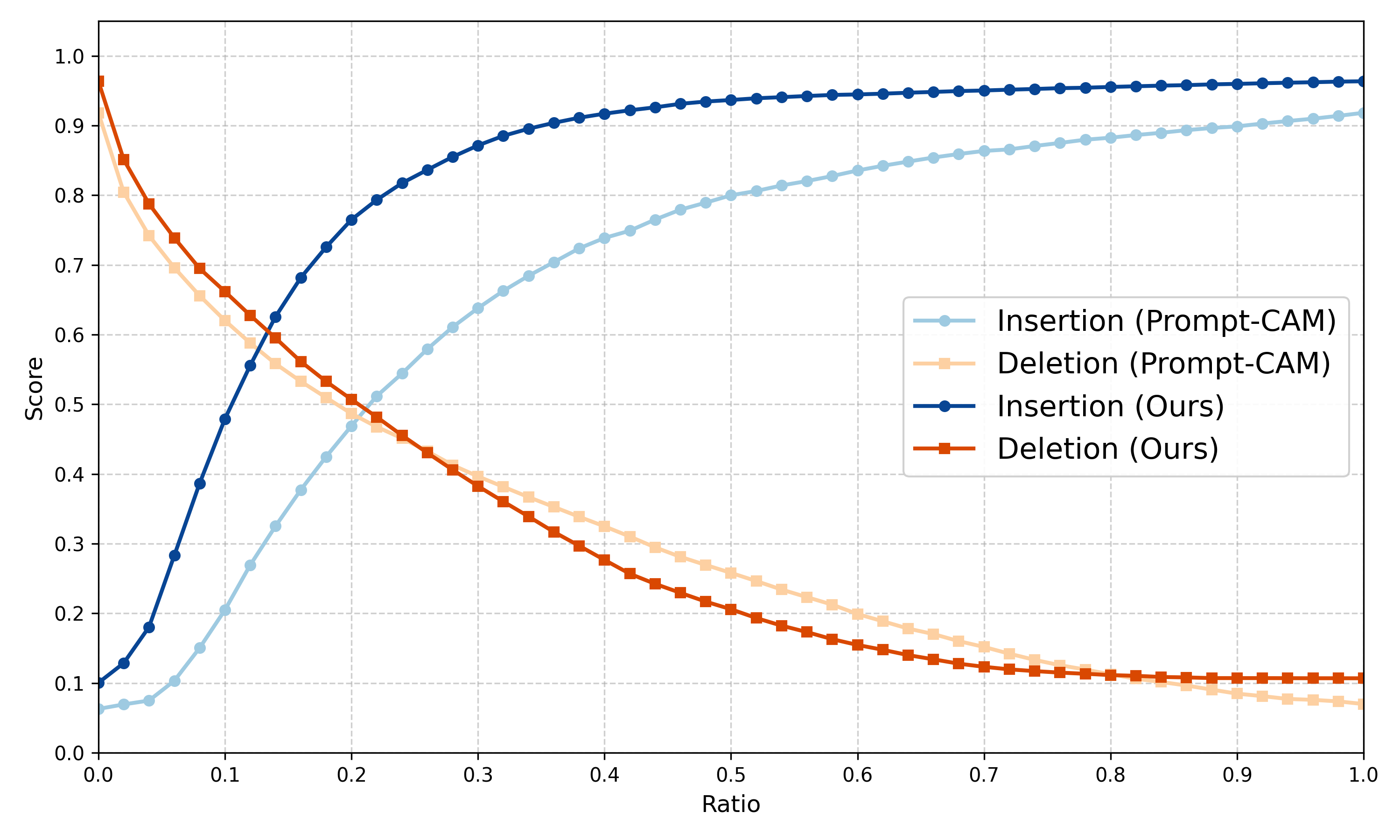}
        \vspace{-0.4em}\centerline{\small (c) Stanford Dogs}
    \end{minipage}
    \caption{Insertion and deletion curve comparisons on CUB, Oxford-IIIT Pet, and Stanford Dogs. The CUB curves include Prompt-CAM, Prompt-CAM with diversity and sparsity attention regularization, and \method{}. The Oxford-IIIT Pet and Stanford Dogs curves include Prompt-CAM and \method{}.}
    \label{fig:insertion_deletion_curves}
\end{figure*}

Under the intrinsic prompt-attention setting, \method{} improves insertion over Prompt-CAM on CUB, Oxford-IIIT Pet, and Stanford Dogs.
This suggests that the top-ranked attention regions selected by \method{} contain more prediction-relevant evidence during the insertion process.
On CUB, the regularization-only variant also improves insertion, indicating that diversity and sparsity constraints can improve the quality of selected-head evidence.
Deletion results are more mixed: \method{} is slightly worse than Prompt-CAM on CUB, but performs better on Oxford-IIIT Pet and Stanford Dogs.
The corresponding insertion and deletion curves are shown in Fig.~\ref{fig:insertion_deletion_curves}.

\begin{table}[t]
\centering
\caption{
Insertion/deletion faithfulness evaluation under post-hoc and class-specific prompt-attention explanation settings.
Prompt-CAM + Reg. denotes the student-only CARE variant with $\mathcal{L}_{\mathrm{div}}+\mathcal{L}_{\mathrm{sp}}$, without the Auxiliary Query Teacher or KD.
}
\label{tab:faithfulness_combined}

\scriptsize
\setlength{\tabcolsep}{2pt}
\renewcommand{\arraystretch}{1.08}

\begin{minipage}[t]{0.29\textwidth}
\centering
{\normalfont (a) Post-hoc methods on CUB}\\[0.35em]
\begin{tabular}{@{}lcc@{}}
\toprule
Method & Ins.$\uparrow$ & Del.$\downarrow$ \\
\midrule
Grad-CAM & 0.61 & \textbf{0.17} \\
Layer-CAM & \textbf{0.77} & 0.28 \\
Att. Rollout & 0.73 & 0.28 \\
\bottomrule
\end{tabular}
\end{minipage}
\hspace{0.015\textwidth}
\begin{minipage}[t]{0.675\textwidth}
\centering
{\normalfont (b) Class-specific prompt-attention methods}\\[0.35em]
\begin{tabular}{@{}lcccccc@{}}
\toprule
\multirow{2}{*}{Method}
& \multicolumn{2}{c}{CUB}
& \multicolumn{2}{c}{Pet}
& \multicolumn{2}{c}{Dog} \\
\cmidrule(lr){2-3}\cmidrule(lr){4-5}\cmidrule(l){6-7}
& Ins.$\uparrow$ & Del.$\downarrow$
& Ins.$\uparrow$ & Del.$\downarrow$
& Ins.$\uparrow$ & Del.$\downarrow$ \\
\midrule
Prompt-CAM
& 0.78 & \textbf{0.13}
& 0.79 & 0.40
& 0.68 & 0.31 \\
Prompt-CAM + Reg.
& 0.85 & 0.15
& -- & --
& -- & -- \\
\method{}
& \textbf{0.86} & 0.16
& \textbf{0.86} & \textbf{0.39}
& \textbf{0.83} & \textbf{0.29} \\
\bottomrule
\end{tabular}
\end{minipage}

\end{table}

\begin{table}[t]
  \centering
    \caption{
    Component ablation study of \method{} on the CUB dataset.
    The teacher branch includes $\mathcal{L}_{\mathrm{aux}}$ by default.
    CARE w/o Teacher/KD removes both the Auxiliary Query Teacher and KD, leaving only attention regularization on the student.
    }
  \label{tab:ablation}
  \begin{tabular}{lcccc}
    \toprule
    Setting & Teacher & KD & Reg. & Top-1 \\
    \midrule
    Prompt-CAM 
    & -- & -- & -- & 74.1 \\
    CARE w/o Reg. 
    & \checkmark & \checkmark & -- & 78.3 \\
    CARE w/o Teacher/KD
    & -- & -- & \checkmark & 73.4 \\
    \method{} 
    & \checkmark & \checkmark & \checkmark & \textbf{78.5} \\
    \bottomrule
  \end{tabular}
\end{table}

\subsection{Ablation Study}

\paragraph{Component ablation.}
The component ablation on CUB examines how each part of \method{} contributes to the final performance, including the Auxiliary Query Teacher, logit-standardized KD, and attention regularization, as reported in Table~\ref{tab:ablation}.
The teacher-distillation variant improves the Prompt-CAM baseline from 74.1\% to 78.3\%, indicating that multi-level teacher supervision is the main source of the accuracy gain.
In contrast, applying $\mathcal{L}_{\mathrm{div}}$ and $\mathcal{L}_{\mathrm{sp}}$ without teacher supervision reduces accuracy to 73.4\%, suggesting that attention regularization alone can over-constrain the class-specific attention pathway.
When teacher-guided distillation and attention regularization are combined, accuracy reaches 78.5\%, outperforming both the Prompt-CAM baseline and the teacher-only variant.
This result suggests that multi-level teacher supervision and attention regularization play complementary roles in improving the Constrained Attention Student.

\begin{figure}[t]
    \centering
    \begin{minipage}[t]{0.48\textwidth}
        \centering
        \includegraphics[width=\linewidth]{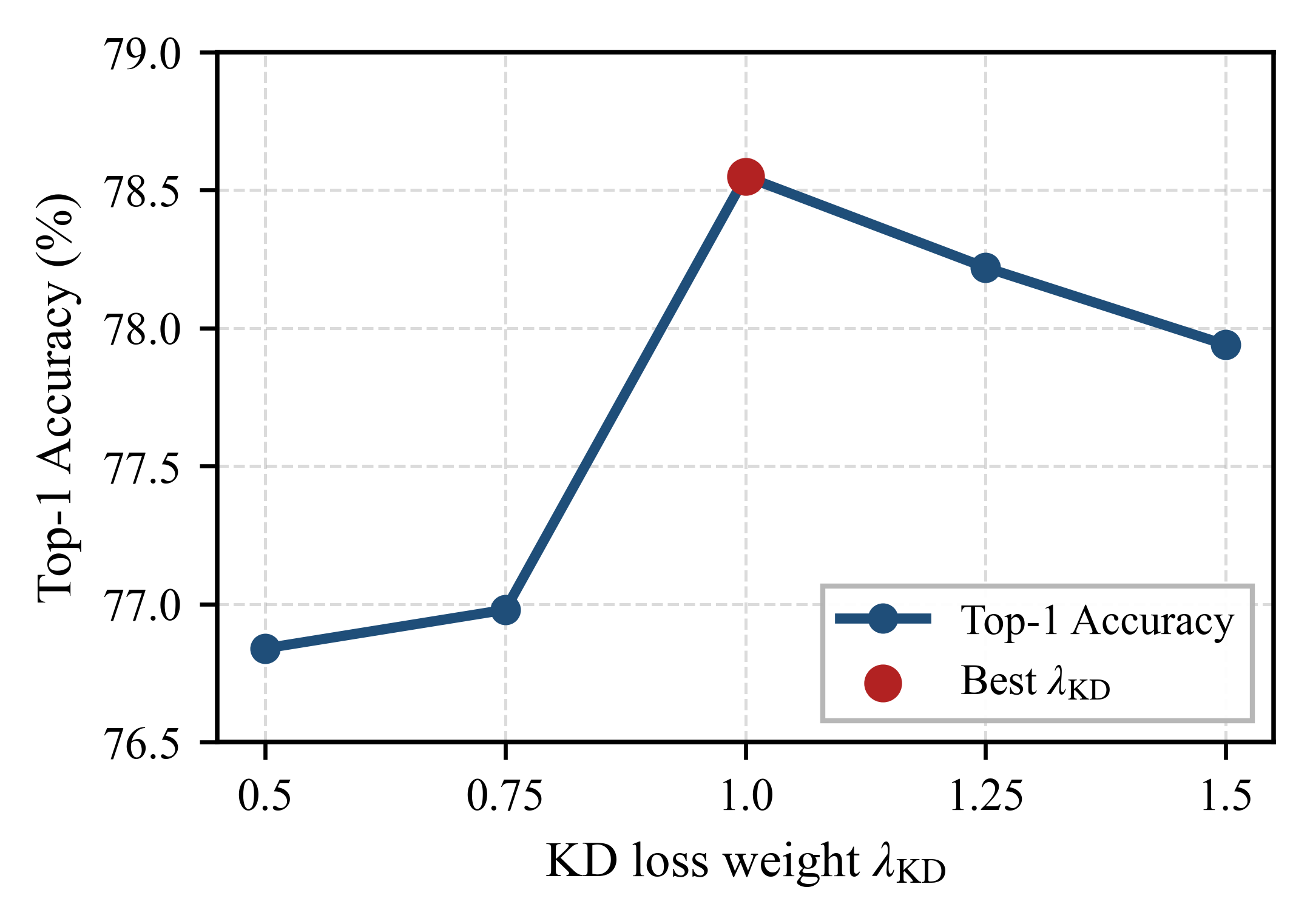}
        \vspace{-0.4em}
        \centerline{\small (a) Classification accuracy}
    \end{minipage}
    \hfill
    \begin{minipage}[t]{0.48\textwidth}
        \centering
        \includegraphics[width=\linewidth]{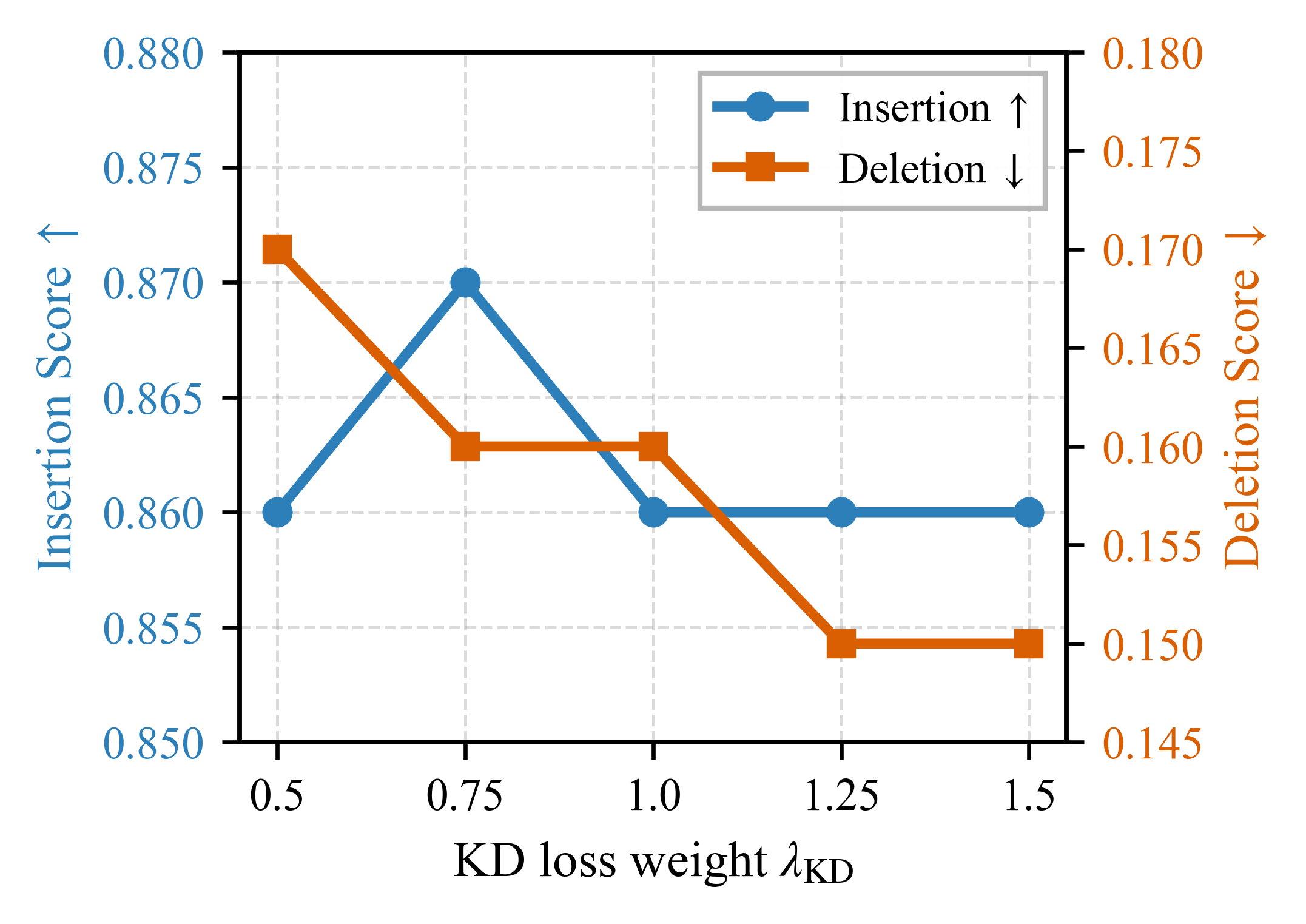}
        \vspace{-0.4em}
        \centerline{\small (b) Faithfulness scores}
    \end{minipage}

    \caption{
    Sensitivity analysis of the KD loss weight $\lambda_{\mathrm{KD}}$ on CUB.
    (a) reports Top-1 classification accuracy.
    (b) reports insertion and deletion faithfulness scores.
    All other hyperparameters are kept identical to the default \method{} configuration.
    }
    \label{fig:kd_weight_sensitivity}
\end{figure}

\paragraph{KD-weight sensitivity.}
Although the framework contains several loss weights, we focus the sensitivity analysis on $\lambda_{\mathrm{KD}}$ because it directly controls the strength of knowledge transfer from the Auxiliary Query Teacher to the Constrained Attention Student.
The accuracy curve in Fig.~\ref{fig:kd_weight_sensitivity}(a) shows that the best result is obtained at $\lambda_{\mathrm{KD}}=1.0$, reaching 78.55\% Top-1 accuracy on CUB.
The faithfulness curves in Fig.~\ref{fig:kd_weight_sensitivity}(b) show that insertion and deletion scores vary only slightly across the tested KD weights.
We therefore use $\lambda_{\mathrm{KD}}=1.0$ as the default setting in the main experiments.

\begin{figure*}[t]
    \centering
    \setlength{\tabcolsep}{2pt}
    \renewcommand{\arraystretch}{1.03}

    \newcommand{\methodtitle}[1]{{\small\bfseries #1}}
    \newcommand{\rowlabel}[1]{{\scriptsize\normalfont #1}}

    \newcommand{\camheader}{%
    \begin{tabular}{@{}*{5}{>{\centering\arraybackslash}p{0.071\textwidth}}@{}}
        \scriptsize Input & \scriptsize Top-1 & \scriptsize Top-2 & \scriptsize Top-3 & \scriptsize Top-4
    \end{tabular}
    }

    \begin{tabular}{
        >{\centering\arraybackslash}m{0.065\textwidth}
        >{\centering\arraybackslash}m{0.445\textwidth}
        >{\centering\arraybackslash}m{0.445\textwidth}
    }
        & \methodtitle{Prompt-CAM} & \methodtitle{CARE} \\[-0.15em]
        & \camheader & \camheader \\[2pt]

        \rowlabel{CUB} &
        \includegraphics[width=0.44\textwidth]{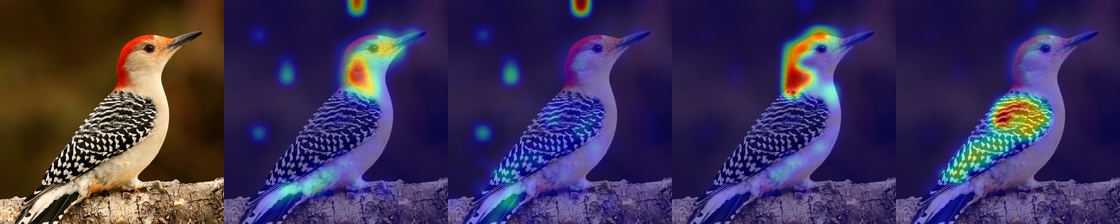} &
        \includegraphics[width=0.44\textwidth]{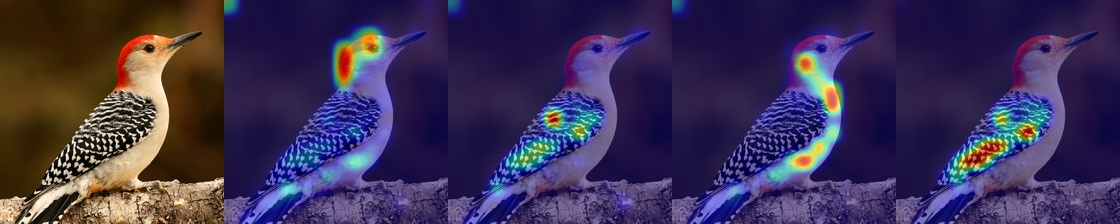} \\[2pt]

        \rowlabel{Pet} &
        \includegraphics[width=0.44\textwidth]{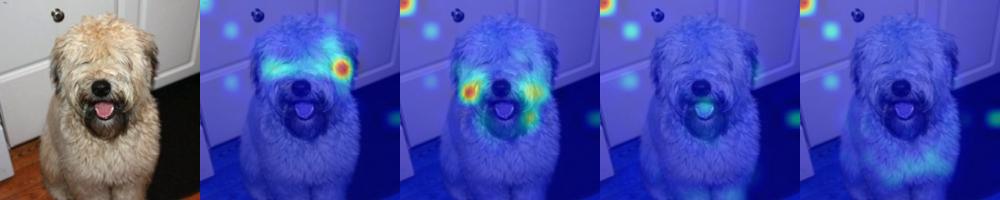} &
        \includegraphics[width=0.44\textwidth]{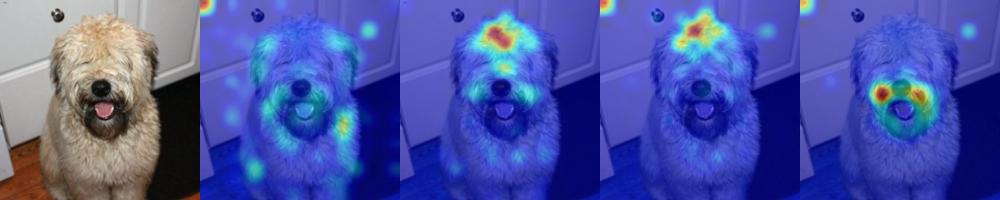} \\[2pt]

        \rowlabel{Dog} &
        \includegraphics[width=0.44\textwidth]{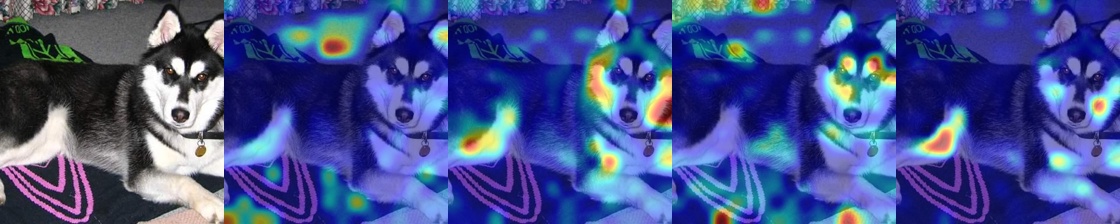} &
        \includegraphics[width=0.44\textwidth]{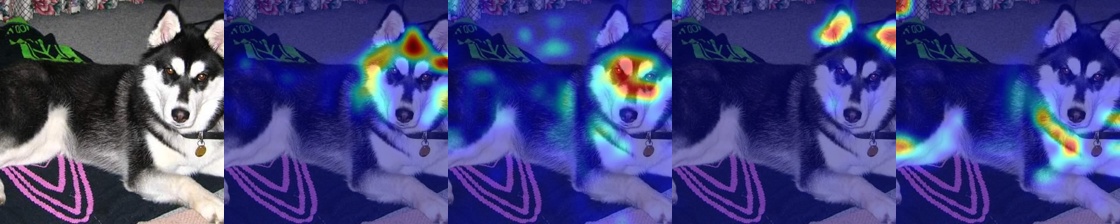} \\[2pt]

        \rowlabel{Car} &
        \includegraphics[width=0.44\textwidth]{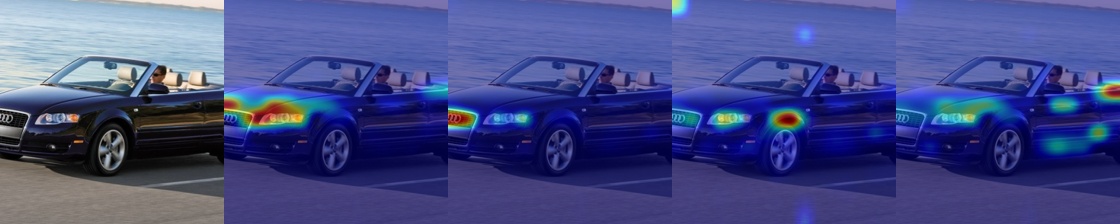} &
        \includegraphics[width=0.44\textwidth]{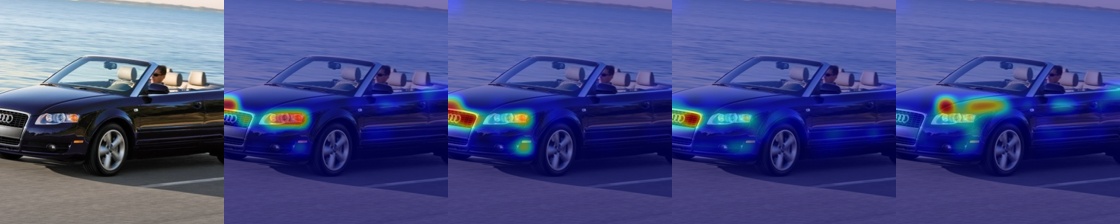} \\
    \end{tabular}

    \caption{
    Visualization of class-specific attention maps from the top-4 attention heads on CUB, Oxford-IIIT Pet, Stanford Dogs, and Stanford Cars.
    }
    \label{fig:qualitative_heatmaps}
\end{figure*}

\begin{table}[t]
  \centering

  \begin{minipage}[t]{0.63\textwidth}
    \centering
    \caption{
    Query-token analysis on the CUB dataset.
    }
    \label{tab:query_token_analysis}
    \scriptsize
    \setlength{\tabcolsep}{3pt}
    \renewcommand{\arraystretch}{1.08}
    \begin{tabular}{lcccc}
      \toprule
      Setting & $N_q$ & Top-1 & Train. Params & Total Params \\
      \midrule
      Prompt-CAM & -- & 74.1 & 1.84M & 87.56M \\
      \method{} & 4  & 77.1 & 25.05M & 110.77M \\
      \method{} & 8  & \textbf{78.5} & 25.06M & 110.78M \\
      \method{} & 16 & 76.7 & 25.07M & 110.79M \\
      \bottomrule
    \end{tabular}
  \end{minipage}
  \hfill
  \begin{minipage}[t]{0.33\textwidth}
    \centering
    \caption{
    Layer combination study on the CUB dataset.
    }
    \label{tab:teacher_layers}
    \scriptsize
    \setlength{\tabcolsep}{5pt}
    \renewcommand{\arraystretch}{1.08}
    \begin{tabular}{lc}
      \toprule
      Teacher indices & Top-1 \\
      \midrule
      4, 8, 11 & \textbf{78.5} \\
      5, 8, 11 & 76.9 \\
      4, 7, 11 & 76.7 \\
      4, 9, 11 & 76.6 \\
      3, 7, 10 & 77.1 \\
      \bottomrule
    \end{tabular}
  \end{minipage}

\end{table}

\subsection{Query Token Analysis}

The number of learnable query tokens controls the readout capacity of the Auxiliary Query Teacher for multi-level feature aggregation.
We analyze this factor on CUB by fixing the selected teacher-layer indices to $\{4,8,11\}$ and changing only $N_q$, with Prompt-CAM included as the baseline without teacher query tokens. 
The results are reported in Table~\ref{tab:query_token_analysis}.

With $N_q=4$, \method{} reaches 77.1\% Top-1 accuracy, improving over Prompt-CAM but remaining below the default setting. 
This suggests that four query tokens may provide insufficient capacity to summarize intermediate representations. 
Increasing $N_q$ to 8 raises accuracy to 78.5\%, indicating that moderate query-token capacity helps the teacher extract multi-level evidence. 
Increasing $N_q$ to 16 lowers accuracy to 76.7\%, suggesting that excessive query tokens may introduce redundant or unstable readouts for distillation.

The parameter counts change only marginally across different values of $N_q$, so the accuracy differences mainly reflect teacher readout capacity rather than parameter scaling. 
We therefore use $N_q=8$ as the default setting.

\subsection{Selecting Complementary Teacher Layers}

The selected teacher-layer indices control the depth distribution of the multi-level supervision provided by the Auxiliary Query Teacher.
As reported in Table~\ref{tab:teacher_layers}, the zero-based index combination $\{4,8,11\}$ achieves the best accuracy, while changing the early, middle, or late layer reduces performance. 
This suggests that the teacher branch benefits from complementary representations across different Transformer depths.

This pattern is consistent with the hierarchy of ViT features: earlier layers preserve local appearance cues, middle layers provide part-level structure, and deeper layers encode semantic information. 
The $\{4,8,11\}$ combination therefore provides a more balanced teacher signal than alternatives at similar depths.

\subsection{Qualitative Discussion}

Qualitative visualizations are used to examine whether the refined class-specific attention maps highlight meaningful fine-grained regions across CUB, Oxford-IIIT Pet, Stanford Dogs, and Stanford Cars, as shown in Fig.~\ref{fig:qualitative_heatmaps}. 
Prompt-CAM and \method{} follow the same visualization protocol for selecting the displayed attention heads. 
Specifically, we adopt the iterative head-pruning procedure used by the Prompt-CAM baseline: at each step, one candidate head is replaced by a uniform attention pattern, the model is re-evaluated, and the head whose removal causes the smallest decrease in target-class confidence is pruned. 
The four remaining heads are visualized as trait-level attention maps. 
This visualization-time selection is separate from the training-time sparsity regularization in Sec.~\ref{sec:attention_reg}.

The selected heads generally correspond to semantically meaningful fine-grained regions, such as the head, beak, neck, ears, or other object-specific local parts. 
Since the heads are selected according to their contribution to target-class confidence rather than human visual preference, some maps may still contain contextual or diffuse responses. 
Compared with Prompt-CAM, \method{} produces more compact and less redundant attention patterns in many examples. 
This qualitative behavior is consistent with the insertion/deletion results and supports the role of attention regularization in refining class-specific localization.

\section{Limitations and Discussion}

\method{} has two main limitations. First, the Auxiliary Query Teacher introduces additional training-time parameters and computational overhead, mainly from the learnable query tokens, multi-layer query readout, feature fusion, and teacher classifier. However, the teacher branch is discarded during inference, so the additional cost does not affect the deployed student model. Second, our experiments are conducted primarily on natural fine-grained datasets. Applying \method{} to domains such as medical or industrial imagery requires further evaluation under domain-specific variations, robustness requirements, and reliability constraints.

\section{Conclusion}

We presented \method{}, a constrained attention refinement framework for faithful fine-grained visual classification. \method{} uses a Constrained Attention Student and a training-only Auxiliary Query Teacher to transfer multi-level discriminative knowledge while preserving class-specific attention as the prediction and explanation source. Experiments on four fine-grained benchmarks show consistent accuracy gains over the class-specific prompt-attention baseline and competitive faithfulness results, indicating that teacher-student distillation and attention regularization can strengthen the interpretable attention pathway without replacing its intrinsic localization mechanism.

\section*{Acknowledgements}
This work was supported in part by the National Natural Science Foundation of China (No.~62306075) and the Natural Science Foundation of Shandong Province (No.~ZR2026QC0725).

\bibliographystyle{splncs04}
\bibliography{references}
\end{document}